\documentclass[runningheads]{llncs}
\usepackage{graphicx}
\usepackage{amsmath,amssymb} 
\usepackage{color}

\usepackage{color,xcolor}
\usepackage{epsfig}
\usepackage{graphicx}

\usepackage{adjustbox}
\usepackage{array}
\usepackage{booktabs}
\usepackage{colortbl}
\usepackage{float,wrapfig}
\usepackage{hhline}
\usepackage{multirow}
\usepackage{subcaption} 
\usepackage[labelfont={bf},labelsep={period},font={small}]{caption}

\let\llncssubparagraph\subparagraph
\let\subparagraph\paragraph
\usepackage[compact]{titlesec}
\let\subparagraph\llncssubparagraph

\usepackage{amsmath,amsfonts,amssymb}
\usepackage{bm}
\usepackage{nicefrac}
\usepackage{microtype}

\usepackage{changepage}
\usepackage{extramarks}
\usepackage{fancyhdr}
\usepackage{lastpage}
\usepackage{setspace}
\usepackage{soul}
\usepackage{xspace}

\usepackage{url}

\usepackage{algorithm, algorithmic}
\usepackage{enumerate}

\newcolumntype{L}[1]{>{\raggedright\let\newline\\\arraybackslash\hspace{0pt}}m{#1}}
\newcolumntype{C}[1]{>{\centering\let\newline\\\arraybackslash\hspace{0pt}}m{#1}}
\newcolumntype{R}[1]{>{\raggedleft\let\newline\\\arraybackslash\hspace{0pt}}m{#1}}

\newcommand{\sects}[1]{Sections~\ref{#1}}
\newcommand{\eqn}[1]{Equation~\ref{#1}}
\newcommand{\fig}[1]{Figure~\ref{#1}}
\newcommand{\figs}[1]{Figures~\ref{#1}}
\newcommand{\tbl}[1]{Table~\ref{#1}}
\newcommand{\tbls}[1]{Tables~\ref{#1}}

\newcommand{\ignore}[1]{}

\makeatletter
\DeclareRobustCommand\onedot{\futurelet\@let@token\@onedot}
\def\@onedot{\ifx\@let@token.\else.\null\fi\xspace}

\def\eg{\emph{e.g}\onedot} 
\def\ie{\emph{i.e}\onedot}

\def\etal{\emph{et al}\onedot}
\makeatother

\definecolor{MyDarkBlue}{rgb}{0,0.08,1}
\definecolor{MyDarkGreen}{rgb}{0.02,0.6,0.02}
\definecolor{MyDarkRed}{rgb}{0.8,0.02,0.02}
\definecolor{MyDarkOrange}{rgb}{0.40,0.2,0.02}
\definecolor{MyPurple}{RGB}{111,0,255}
\definecolor{MyRed}{rgb}{1.0,0.0,0.0}
\definecolor{MyGold}{rgb}{0.75,0.6,0.12}
\definecolor{MyDarkgray}{rgb}{0.66, 0.66, 0.66}

\newcommand{\myparagraph}[1]{\noindent{\bf #1}}

\newcommand{\model}{ShapeHD\xspace}

\begin{document}
\title{Learning Shape Priors for \\Single-View 3D Completion and Reconstruction}

\titlerunning{Learning Shape Priors for Single-View 3D Completion and Reconstruction}
%
\author{Jiajun Wu*\inst{1} \and
Chengkai Zhang*\inst{1} \and
Xiuming Zhang\inst{1} \and
Zhoutong Zhang\inst{1} \and \\
William T. Freeman\inst{1,2} \and
Joshua B. Tenenbaum\inst{1}}
%
\authorrunning{J. Wu et al.}
%

\institute{MIT CSAIL, Cambridge MA 02139, USA \and
Google Research, Cambridge MA 02139, USA}
\footnotetext{* J. Wu and C. Zhang contributed equally to this work.}

\maketitle              
\begin{figure}[t]
    \centering
    \includegraphics[width=\linewidth]{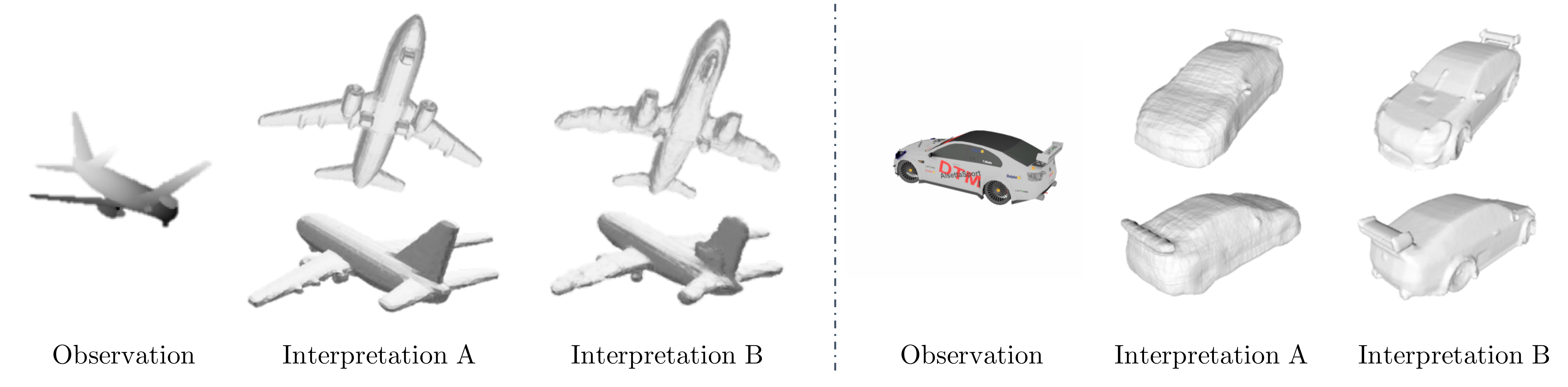}
    \vspace{-18pt}
    \caption{
    Our model completes or reconstructs the object's full 3D shape with fine details from a single depth or RGB image. In this figure, we show two examples, each consisting of an input image, two views of its ground truth shape, and two views of our results. Our reconstructions are of high quality with fine details, and are preferred by humans 41\% and 35\% of the time in behavioral studies, respectively. Our model takes a single feed-forward pass without any post-processing during testing, and is thus highly efficient ($<100$ ms) and practically useful. Answers are available in the footnote.} 
    \vspace{-18pt}
    \label{fig:teaser} 
\end{figure}

\begin{abstract}

The problem of single-view 3D shape completion or reconstruction is challenging, because among the many possible shapes that explain an observation, most are implausible and do not correspond to natural objects. Recent research in the field has tackled this problem by exploiting the expressiveness of deep convolutional networks. In fact, there is another level of ambiguity that is often overlooked: among plausible shapes, there are still multiple shapes that fit the 2D image equally well; \ie, the ground truth shape is non-deterministic given a single-view input. Existing fully supervised approaches fail to address this issue, and often produce blurry mean shapes with smooth surfaces but no fine details.

In this paper, we propose \emph{\model}, pushing the limit of single-view shape completion and reconstruction by integrating deep generative models with adversarially learned shape priors. The learned priors serve as a regularizer, penalizing the model only if its output is unrealistic, not if it deviates from the ground truth. Our design thus overcomes both levels of ambiguity aforementioned. Experiments demonstrate that \model outperforms state of the art by a large margin in both shape completion and shape reconstruction on multiple real datasets.


\end{abstract}

\vspace{-10pt}
\section{Introduction}
\label{sec:intro}

Let's start with a game: each of the two instances in \fig{fig:teaser} shows a depth or color image and two different 3D shape interpretations. Which one looks better?

We asked this question to 100 people on Amazon Mechanical Turk. 59\% of them preferred interpretation A of the airplane, and 35\% preferred interpretation A of the car. These numbers suggest that people's opinions diverge on these two cases, indicating that these reconstructions are close in quality, and their perceptual differences are relatively minor. 

Actually, for each instance, one of the reconstructions is the output of the model introduced in this paper, and the other is the ground truth shape. Answers are available in the footnote.

\begin{figure}[t]
\includegraphics[width=\linewidth]{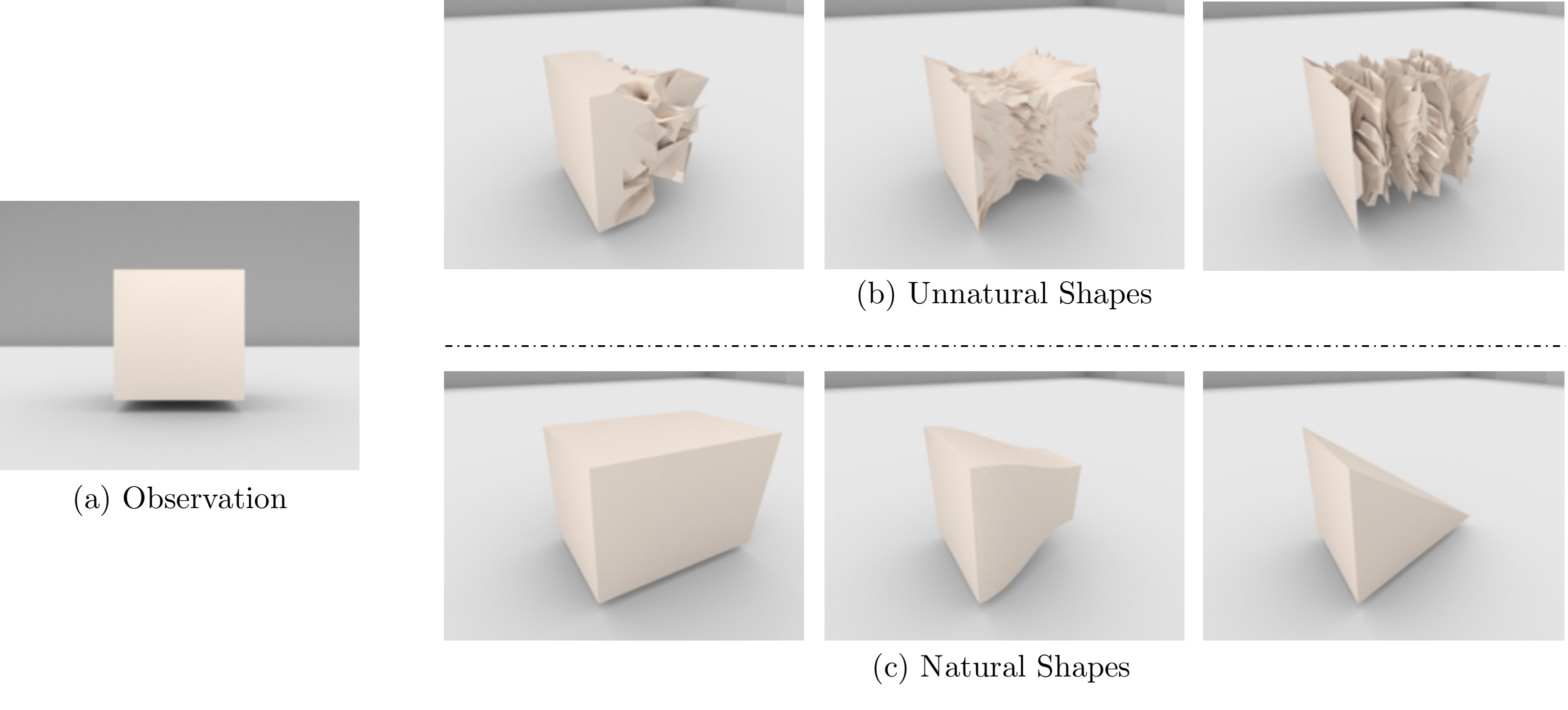}
\vspace{-24pt}
\caption{Two levels of ambiguity in single-view 3D shape perception. For each 2D observation (a), there exist many possible 3D shapes that explain this observation equally well (b, c), but only a small fraction of them correspond to real, daily shapes (c). Methods that exploit deep networks for recognition reduce, to a certain extent, ambiguity on this level. By using an adversarially learned naturalness model, our \model aims to model ambiguity on the next level: even among the realistic shapes, there are still multiple shapes explaining the observation well (c).}
\label{fig:illposed}
\vspace{-22pt}
\end{figure}

In this paper, we aim to push the limits of 3D shape completion from a single depth image, and of 3D shape reconstruction from a single color image. Recently, researchers have made impressive progress on the these tasks~\cite{Choy20163d,Tulsiani2017Multi,Dai2017Shape}, making use of gigantic 3D datasets~\cite{Chang2015Shapenet:,Xiang2014PASCAL:,Xiang2016Objectnet3d:}. Many of these methods tackle the ill-posed nature of the problem by using deep convolutional networks to regress possible 3D shapes. Leveraging the power of deep generative models, their systems learn to avoid producing implausible shapes (\fig{fig:illposed}b).

However, from \fig{fig:illposed}c we realize that there is still ambiguity that a supervisedly trained network fails to model. From just a single view, there exist multiple natural shapes that explain the observation equally well. In other words, there is no deterministic ground truth for each observation. Through pure supervised learning, the network tends to generate mean shapes that minimize its penalty precisely due to this ambiguity.

To tackle this, we propose \model, which completes or reconstructs a 3D shape by combining deep volumetric convolutional networks with adversarially learned shape priors. The learned shape priors penalize the model only if the generated shape is unrealistic, not if it deviates from the ground truth. This overcomes the difficulty discussed above. Our model characterizes this naturalness loss through adversarial learning, a research topic that has received immense attention in recent years and is still rapidly growing~\cite{Goodfellow2014Generative,Radford2016Unsupervised,Wu2016Learning}.

Experiments on multiple synthetic and real datasets suggest that \model performs well on single-view 3D shape completion and reconstruction, achieving better results than state-of-the-art systems. Further analyses reveal that the network learns to attend to meaningful object parts, and the naturalness module indeed helps to characterize shape details over time. 

\footnotetext{\raisebox{\depth}{\rotatebox{180}{Our reconstructions: B, A}}}

\section{Related Work}
\label{sec:related}

\myparagraph{3D shape completion. }
Shape completion is an essential task in geometry processing and has wide applications. Traditional methods have attempted to complete shapes with local surface primitives, or to formulate it as an optimization problem~\cite{nealen2006laplacian,sorkine2004least}, \eg, Poisson surface reconstruction solves an indicator function on a voxel grid via the
Poisson equation~\cite{kazhdan2013screened,Kazhdan2006Poisson}. Recently, there have also been a growing number of papers on exploiting shape structures and regularities~\cite{Mitra2006Partial,thrun2005shape}, and papers on leveraging strong database priors~\cite{Sung2015Data,li2015database,brock2016generative}. These methods, however, often require the database to contain exact parts of the shape, and thus have limited generalization power.

With the advances in large-scale shape repositories like ShapeNet~\cite{Chang2015Shapenet:}, researchers began to develop fully data-driven methods, some building upon deep convolutional networks. To name a few, Voxlets~\cite{firman-cvpr-2016} employs random forests for predicting unknown voxel neighborhoods. 3D ShapeNets~\cite{Wu20153d} uses a deep belief network to obtain a generative model for a given shape database, and Nguyen~\etal~\cite{thanh2016field} extend the method for mesh repairing. 

Probably the most related paper to ours is the 3D-EPN from Dai~\etal~\cite{Dai2017Shape}. 3D-EPN achieves impressive results on 3D shape completion from partial depth scans by levering 3D convolutional networks and nonparametric patch-based shape synthesis methods. Our model has advantages over 3D-EPN in two aspects. First, with naturalness losses, \model can choose among multiple hypotheses that explain the observation, therefore reconstructing a high-quality 3D shape with fine details; in contrast, the output from 3D-EPN without nonparametric shape synthesis is often blurry. Second, our completion takes a single feed-forward pass without any post-processing, and is thus much faster ($<$100ms) than 3D-EPN.

\myparagraph{Single-image 3D reconstruction. }
The problem of recovering the object shape from a single image is challenging, as it requires both powerful recognition systems and prior shape knowledge. As an early attempt, Huang~\etal~\cite{Huang2015Single} propose to borrows shape parts from existing CAD models. With the development of large-scale shape repositories like ShapeNet~\cite{Chang2015Shapenet:} and methods like deep convolutional networks, researchers have built more scalable and efficient models in recent years~\cite{Choy20163d,Girdhar2016Learning,Haene2017Hierarchical,Kar2015Category,Novotny2017Learning,Rezende2016Unsupervised,Tatarchenko2016Multi,Tulsiani2017Multi,Wu2017MarrNet:,Wu2016Learning,Yan2016Perspective}. While most of these approaches encode objects in voxels from vision, there have also been attempts to reconstruct objects in point clouds~\cite{Fan2017point,groueix2017} or octave trees~\cite{Riegler2017Octnet:,Tatarchenko2017Octree,Riegler2017OctNetFusion}, or using tactile signals~\cite{touch}. 

A related direction is to estimate 2.5D sketches (\eg, depth and surface normal maps) from an RGB image. In the past, researchers have explored recovering 2.5D sketches from shading, texture, or color images~\cite{Barron2015Shape,Bell2014Intrinsic,Horn1989Shape,Tappen2003Recovering,Weiss2001Deriving,Zhang1999Shape}. With the development of depth sensors~\cite{Izadi2011KinectFusion:} and larger-scale RGB-D datasets~\cite{McCormac2017SceneNet,Silberman2012Indoor,Song2017Semantic}, there have also been papers on estimating depth~\cite{Chen2016Single,Eigen2015Predicting}, surface normals~\cite{Bansal2016Marr,Wang2015Designing}, and other intrinsic images~\cite{janner2017intrinsic,Shi2017Learning} with deep networks. Inspired by MarrNet~\cite{Wu2017MarrNet:}, we reconstructs 3D shapes via modeling 2.5D sketches, but incorporating a naturalness loss for much higher quality. 

\myparagraph{Perceptual losses and adversarial learning. }
Researchers recently proposed to evaluate the quality of 2D images using perceptual losses~\cite{johnson2016perceptual,dosovitskiy2016generating}. The idea has been applied to many image tasks like style transfer and super-resolution~\cite{johnson2016perceptual,ledig2016photo}. Furthermore, the idea has been extended to learn a perceptual loss function with generative adversarial nets (GAN)~\cite{Goodfellow2014Generative}. GANs incorporate an adversarial discriminator into the procedure of generative modeling, and achieve impressive performance on tasks like image synthesis~\cite{Radford2016Unsupervised}. Isola~\etal~\cite{Isola2016Learning} and Zhu~\etal~\cite{Zhu2016Generative} use GANs for image translation with and without supervision, respectively.

In 3D vision, Wu~\etal~\cite{Wu2016Learning} extends GANs for 3D shape synthesis. However, their model for shape reconstruction (3D-VAE-GAN) often produces a noisy, incomplete shape given an RGB image. This is because training GANs jointly with recognition networks could be highly unstable. Many other researchers have also noticed this issue: although adversarial modeling of 3D shape space may resolve the ambiguity discussed earlier, its training could be challenging~\cite{Dai2017Shape}. Addressing this, when Gwak~\etal~\cite{Gwak2017Weakly} explored adversarial nets for single-image 3D reconstruction and chose to use GANs to model 2D projections instead of 3D shapes. This weakly supervised setting, however, hampers their reconstructions. In this paper, we develop our naturalness loss by adversarial modeling of the 3D shape space, outperforming the state-of-the-art significantly.

\section{Approach}
\label{sec:model}

\begin{figure}[t]
\centering
\includegraphics[width=\linewidth]{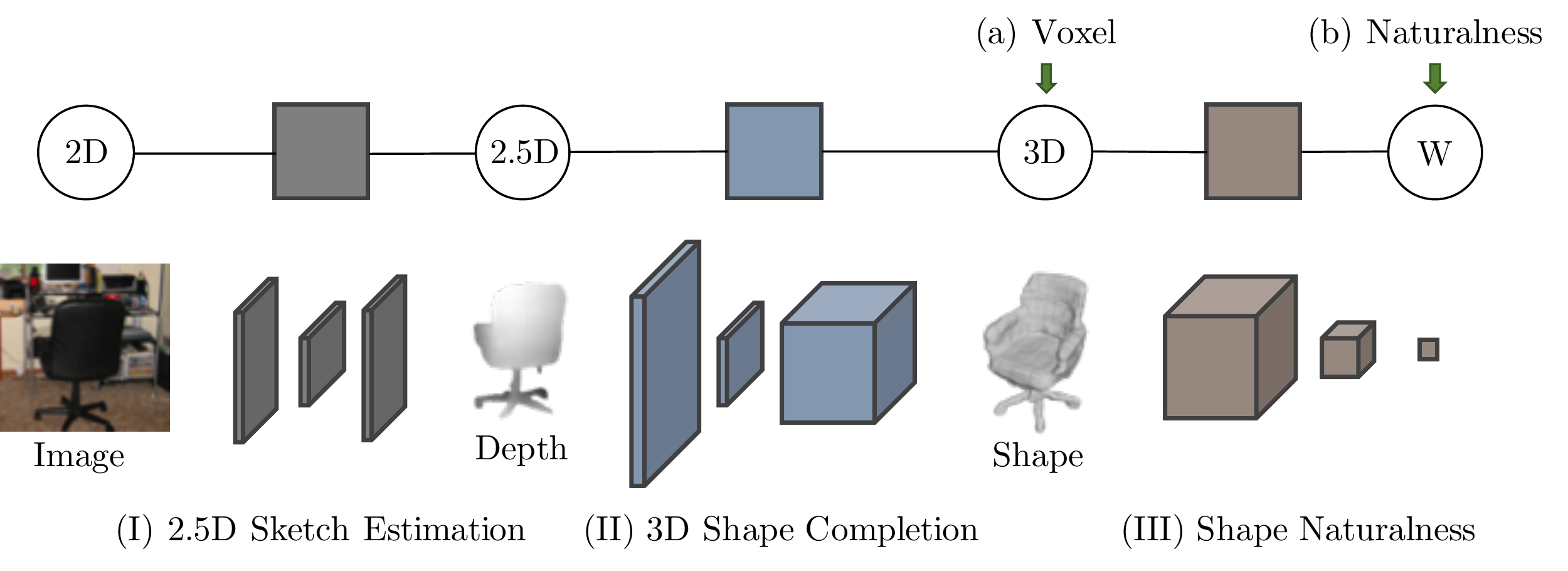}
\vspace{-25pt}
\caption{For single-view shape reconstruction, \model contains three components: (I) a 2.5D sketch estimator that predicts depth, surface normal and silhouette images from a single image; (II) a 3D shape completion module that regresses 3D shapes from silhouette-masked depth and surface normal images; (III) an adversarially pretrained convolutional net that serves as the naturalness loss function. While fine-tuning the 3D shape completion net, we use two losses: a supervised loss on the output shape, and a naturalness loss offered by the pretrained discriminator.}
\label{fig:model}
\vspace{-20pt}
\end{figure}

Our model consists of three components: a 2.5D sketch estimator and a 3D shape estimator that predicts a 3D shape from an RGB image via 2.5D sketches (\fig{fig:model}-I,II, inspired by MarrNet~\cite{Wu2017MarrNet:}), and a deep naturalness model that penalizes the shape estimator if the predicted shape is unnatural (\fig{fig:model}-III). Models trained with a supervised reconstruction loss alone often generate blurry mean shapes. Our learned naturalness model helps to avoid this issue.

\myparagraph{2.5D sketch estimation network. }
Our 2.5D sketch estimator has an encoder-decoder structure that predicts the object's depth, surface normals, and silhouette from an RGB image (\fig{fig:model}-I). We use a ResNet-18~\cite{He2015Deep} to encode a 256$\times$256 image into 512 feature maps of size $8\times 8$. The decoder consists of four transposed convolutional layers with a kernel size of $5\times5$ and a stride and padding of 2. The predicted depth and surface normal images are then masked by the predicted silhouette and used as the input to our shape completion network.

\myparagraph{3D shape completion network. }
Our 3D estimator (\fig{fig:model}-II) is an encoder-decoder network that predicts a 3D shape in the canonical view from 2.5D sketches. The encoder is adapted from ResNet-18~\cite{He2015Deep} to encode a four-channel 256$\times$256 image (one for depth, three for surface normals) into a 200-D latent vector. The vector then goes through a decoder of five transposed convolutional and ReLU layers to generate a 128$\times$128$\times$128 voxelized shape. Binary cross-entropy losses between predicted and target voxels are used as the supervised loss~$L_{\mathrm{voxel}}$.

\subsection{Shape Naturalness Network}
Due to the inherent uncertainty of single-view 3D shape reconstruction, shape completion networks with only a supervised loss usually predict unrealistic mean shapes. By doing so, they minimize the loss when there exist multiple possible ground truth shapes. We instead introduce an adversarially trained deep naturalness regularizer that penalizes the network for such unrealistic shapes. 

We pre-train a 3D generative adversarial network~\cite{Goodfellow2014Generative} to determine whether a shape is realistic. Its generator synthesizes a 3D shape from a randomly sampled vector, and its discriminator distinguishes generated shapes from real ones. Therefore, the discriminator has the ability to model the real shape distribution and can be used as a naturalness loss for the shape completion network. The generator is not involved in our later training process. Following 3D-GAN~\cite{Wu2016Learning}, we use 5 transposed convolutional layers with batch normalization and ReLU for the generator, and 5 convolutional layers with leaky ReLU for the discriminator.

Due to the high dimensionality of 3D shapes (128$\times$128$\times$128), training a GAN becomes highly unstable. To deal with this issue, we follow Gulrajani~\etal~\cite{gulrajani2017improved} and use the Wasserstein GAN loss with a gradient penalty to train our adversarial generative network. Specifically,
\vspace{-5pt}
\begin{equation}
    L_{\mathrm{WGAN}} = \underset{\tilde x\sim P_g}{\mathbb{E}}[D(\tilde x)] - \underset{x\sim P_r}{\mathbb{E}}[D(x)] + \lambda \underset{\hat x\sim P_x}{\mathbb{E}} [(\lVert\triangledown_{\hat x}D(\hat x)\rVert_2 - 1)^2],
    \label{eq:ganloss}
\vspace{-5pt}
\end{equation}
where $D$ is the discriminator, $P_g$ and $P_r$ are distributions of generated shapes and real shapes, respectively. The last term is the gradient penalty from Gulrajani~\etal~\cite{gulrajani2017improved}. During training, the discriminator attempts to minimize the overall loss $L_{\mathrm{WGAN}}$ while the generator attempts to maximize the loss via the first term in~\eqn{eq:ganloss}, so we can define our naturalness loss as
$L_{\mathrm{natural}}=-\underset{\tilde x\sim P_c}{\mathbb{E}}[D(\tilde x)]$,
where $P_c$ are the reconstructed shapes from our completion network.

\subsection{Training Paradigm}

We train our network in two stages. We first pre-train the three components of our model separately. The shape completion network is then fine-tuned with both voxel loss and naturalness losses. 

Our 2.5D sketch estimation network and 3D completion network are trained with images rendered with ShapeNet~\cite{Chang2015Shapenet:} objects (see \sects{sec:data-prep} and \ref{sec:exp_recon} for details). We train the 2.5D sketch estimator using a L2 loss and SGD with a learning rate of 0.001 for 120 epochs. We only use the supervised loss $L_{\mathrm{voxel}}$ for training the 3D estimator at this stage, again with SGD, a learning rate of 0.1, and a momentum of 0.9 for 80 epochs. The naturalness network is trained in an adversarial manner, where we use Adam~\cite{Kingma2015Adam:} with a learning rate of 0.001 and a batch size of 4 for 80 epochs. We set $\lambda=10$ as suggested in Gulrajani~\etal~\cite{gulrajani2017improved}. 

We then fine-tune our completion network with both voxel loss and naturalness losses as $L=L_{\mathrm{voxel}} + \alpha L_{\mathrm{natural}}$. We compare the scale of gradients from the losses and train our completion network with $\alpha=2.75\times 10^{-11}$ using SGD for 80 epochs. Our model is robust to these parameters; they are only for ensuring gradients of various losses are of the same magnitude. 

An alternative is to jointly train the naturalness module with the completion network from scratch using both losses. It seems tempting, but in practice we find that Wasserstein GANs have large losses and gradients, resulting in unstable outputs. We therefore choose to use our pre-training and fine-tuning setup. 

\section{Single-View Shape Completion}
\label{sec:exp}

For 3D shape completion from a single depth image, we only use the last two modules of the model: the 3D shape estimator and deep naturalness network.

\subsection{Setup}
\label{sec:data-prep}

\myparagraph{Data. } We render each of the ShapeNet Core55~\cite{Chang2015Shapenet:} objects from the aeroplane, car and chair categories in 20 random, fully unconstrained views. For each view, we randomly set the azimuth and elevation angles of the camera, but the camera up vector is fixed to be the world $+y$ axis, and the camera always looks at the object center. The focal length is fixed at 50mm with a 35mm film. We use Mitsuba~\cite{Mitsuba}, a physically-based graphics engine, for all our renderings. We used 90\% of the data for training and 10\% for testing.

We render the ground-truth depth image of each object in all 20 views. Depth values are measured from the camera center (\ie, ray depth), rather than from the image plane. To approximate depth scanner data, we also generate the accompanying ground-truth surface normal images from the raw depth data, as surface normal maps are the common by-products of depth scanning. All our rendered surface normal vectors are defined in the camera space.

\myparagraph{Baselines. } We compare with the state of the art: 3D-EPN~\cite{Dai2017Shape}. To ensure a fair comparison, we convert depth maps to partial surfaces registered in a canonical global coordinate defined by ShapeNet Core55~\cite{Chang2015Shapenet:}, which is required by 3D-EPN. While the original 3D-EPN paper generates their partial observations by rendering and fusing multi-view depth maps, our method takes a single-view depth map as input and is solving a more challenging problem. 

\myparagraph{Metrics. } We use two standard metrics for quantitative comparisons: Intersection over Union (IoU) and Chamfer Distance (CD). In particular, Chamfer distance can be applied to various shape representations including voxels (by sampling points on the isosurface) and point clouds.

\begin{figure}[t]
\centering
\includegraphics[width=\linewidth]{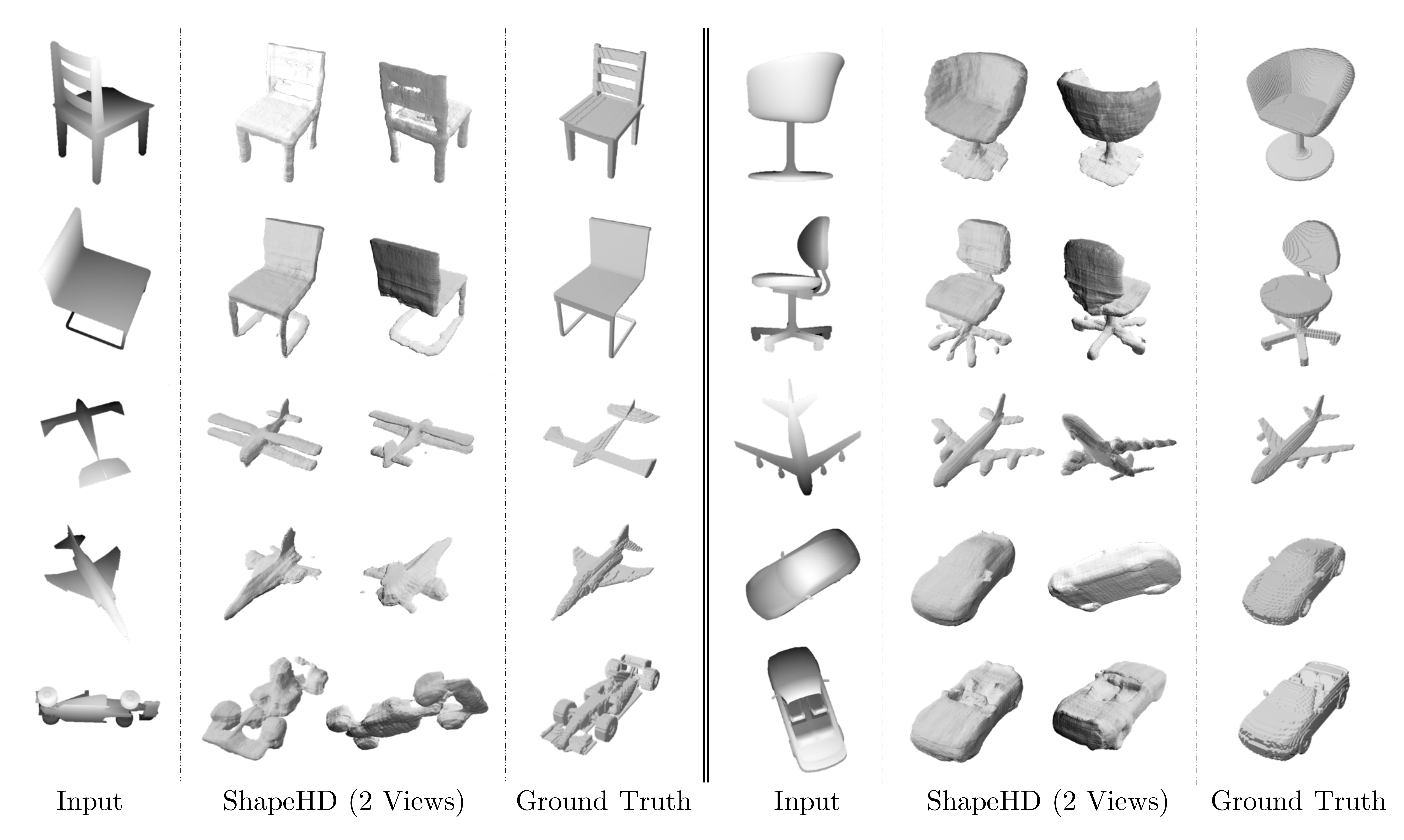}
\vspace{-20pt}
\caption{Results on 3D shape completion from single-view depth. From left to right: input depth maps, shapes reconstructed by \model in the canonical view and a novel view, and ground truth shapes in the canonical view. Assisted by the adversarially learned naturalness losses, \model recovers highly accurate 3D shapes with fine details. Sometimes the reconstructed shape deviates from the ground truth, but can be viewed as another plausible explanation of the input (\eg, the airplane on the left, third row).}
\label{fig:results_comp_shapenet}
\vspace{-20pt}
\end{figure}
\begin{figure}[t!]
\centering
\includegraphics[width=\linewidth]{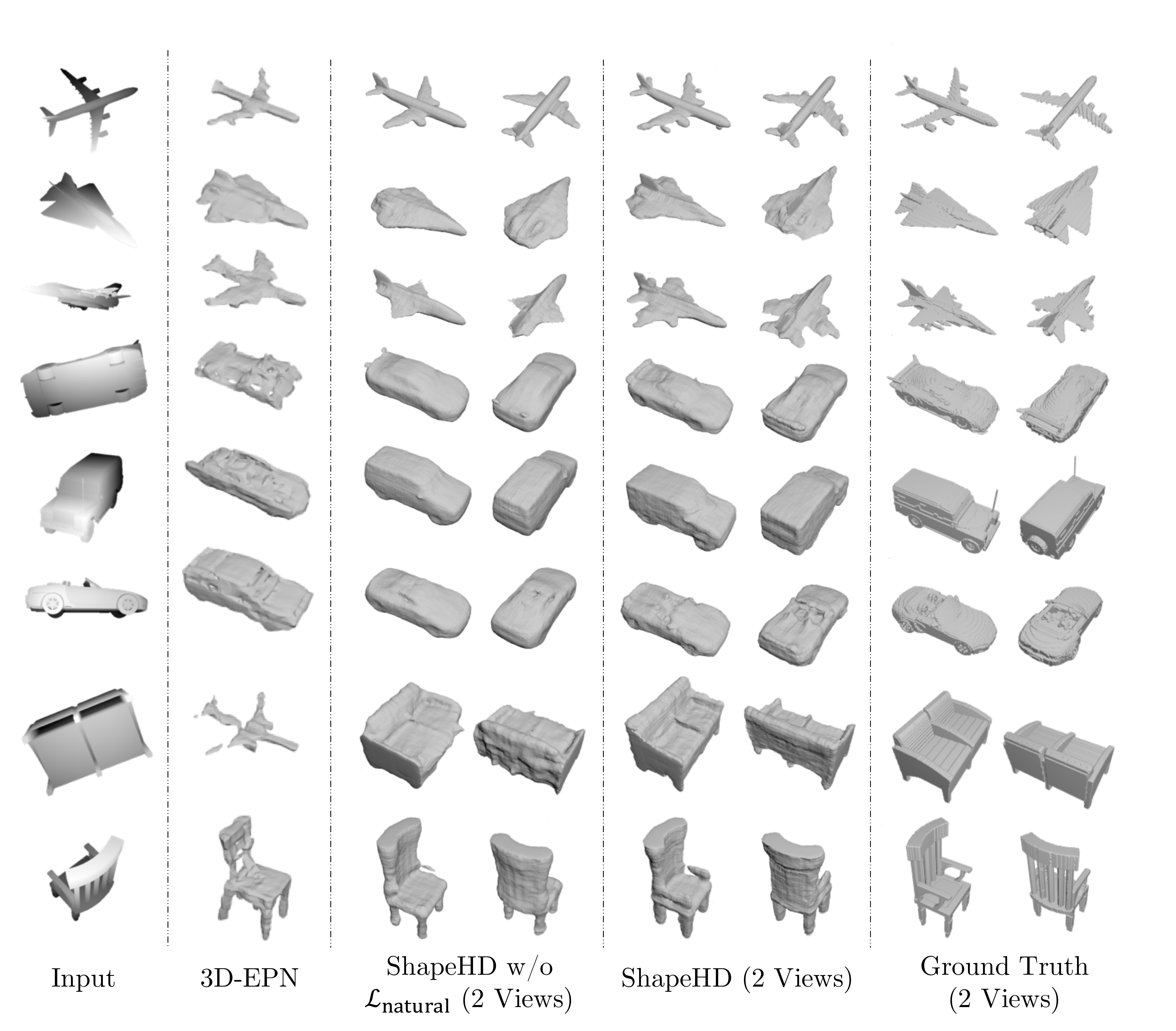}
\vspace{-20pt}
\caption{Our results on 3D shape completion, compared with the state of the art, 3D-EPN~\cite{Dai2017Shape}, and our model but without naturalness losses. Our results contain more details than 3D-EPN. We observe that the adversarially trained naturalness losses help fix errors, add details (\eg, the plane wings in row 3, car seats in row 6, and chair arms in row 8), and smooth planar surfaces (\eg, the sofa back in row 7).}
\label{fig:results_cpr}
\vspace{-10pt}
\end{figure}

\subsection{Results on ShapeNet}

\myparagraph{Qualitative results. } 
In~\fig{fig:results_comp_shapenet}, we show 3D shapes predicted by \model from single-view depth images. While common encoder-decoder structure usually generates mean shapes with few details, our \model predicts shapes with large variance and fine details. In addition, even when there is strong occlusion in the depth image, our model can predict a high-quality, plausible 3D shape that looks good perceptually, and infer parts not present in the input images.  

\myparagraph{Ablation. }
When using naturalness loss, the network is penalized for generating mean shapes that are unreasonable but minimize the supervised loss. In~\fig{fig:results_cpr}, we show reconstructed shapes from our \model with and without naturalness loss (\ie before fine-tuning with $L_{\mathrm{natural}}$), together with ground truth shapes and shapes predicted by 3D-EPN~\cite{Dai2017Shape}. Our results contain finer details compared with those from 3D-EPN. Also, the performance of \model improves greatly with the naturalness loss, which predicts more reasonable and complete shapes.   

\begin{table}[t!]
 	\centering
 	\small
 	\setlength{\tabcolsep}{5pt}
    \begin{tabular}{lcccccccc}
    \toprule
     \multirow{2}{*}{Methods} & \multicolumn{4}{c}{IoU} & \multicolumn{4}{c}{CD} \\
     \cmidrule(r){2-5}\cmidrule(l){6-9}
     & chair & car & plane & avg & chair & car & plane & avg \\
    \midrule
    3D-EPN~\cite{Dai2017Shape} & .147 & .274 & .155 & .181 & .227 & .200 & .125 & .192 \\
    \model w/o $L_{\mathrm{natural}}$ & .466 & {\bf .698} & {\bf .488} & {\bf .529} & .112 & .083 & .071 & .093 \\
    \model & {\bf .488} & {\bf .698} & .452 & {\bf .529} & {\bf .096} & {\bf .078} & {\bf .068} & .{\bf 084} \\
    \bottomrule
    \end{tabular}
    \vspace{5pt}
    \caption{Average IoU scores ($32^3$) and CDs for 3D shape completion on ShapeNet~\cite{Chang2015Shapenet:}. Our model outperforms the state of the art by a large margin. The learned naturalness losses consistently improve the CDs between our results and ground truth. }
    \vspace{-30pt}
    \label{tbl:comp_shapenet}
\end{table}

\begin{figure}[t]
\centering
\includegraphics[width=\linewidth]{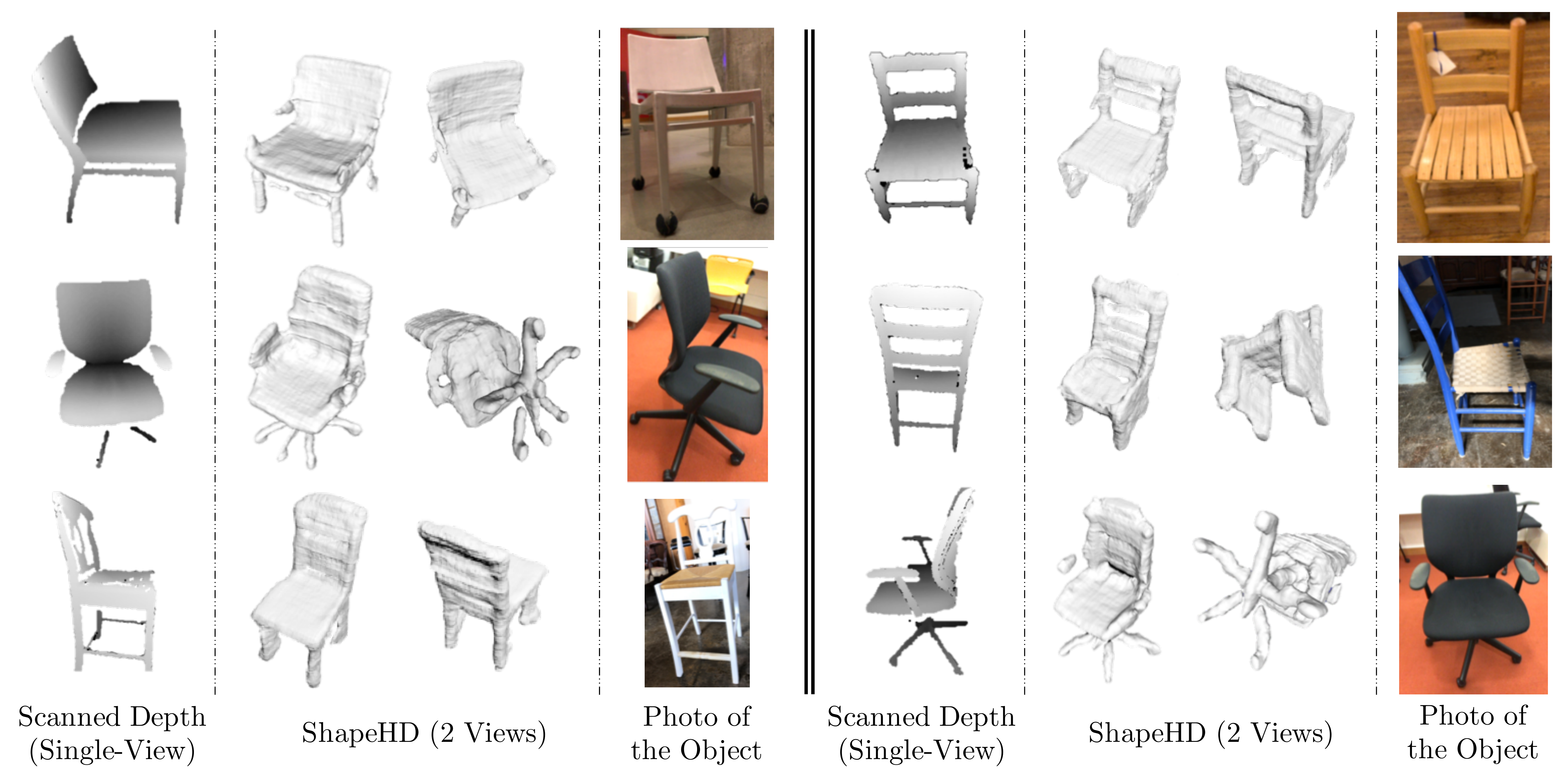}
\vspace{-22pt}
\caption{Results of 3D shape completion on depth data from a physical scanner. Our model is able to reconstruct the shape well from just a single view. From left to right: input depth,  two views of our completion results, and a color image of the object.}
\label{fig:results_comp_pix3d}
\vspace{-20pt}
\end{figure}

\myparagraph{Quantitative results. }
We present quantitative results in \tbl{tbl:comp_shapenet}. Our \model outperforms the state of the art by a margin in all metrics. Our method outputs shapes at the resolution of $128^3$, while shapes produced by 3D-EPN are of resolution $32^3$. Therefore, for a fair comparison, we downsample our predicted shapes to $32^3$ and report results of both methods in that resolution. The original 3D-EPN paper suggests a post-processing step that retrieves similar patches from a shape database for results of a higher resolution. Practically, we find this steps takes 18 hours for a single image. We therefore report results without post-processing for both methods. 

\tbl{tbl:comp_shapenet} also suggests the naturalness loss improve the completion results, achieving comparable IoU scores and better (lower) CDs. CD has been reported to be better at capturing human perception of shape quality~\cite{pix3d}. 

\subsection{Results on Real Depth Scans}
We now show results of \model on real depth scans. We capture six depth maps of different chairs using a Structure sensor (\url{http://structure.io}) and use the captured depth maps to evaluate our model. All the corresponding normal maps used as inputs are estimated from depth measurements. \fig{fig:results_comp_pix3d} shows that \model completes 3D shapes well given a single-view depth map. Our \model is more flexible than 3D-EPN, as we do not need any camera intrinsics or extrinsics to register depth maps. In our case, none of these parameters are known and thus 3D-EPN cannot be applied.

\begin{figure}[t]
 	\centering
    \includegraphics[width=\linewidth]{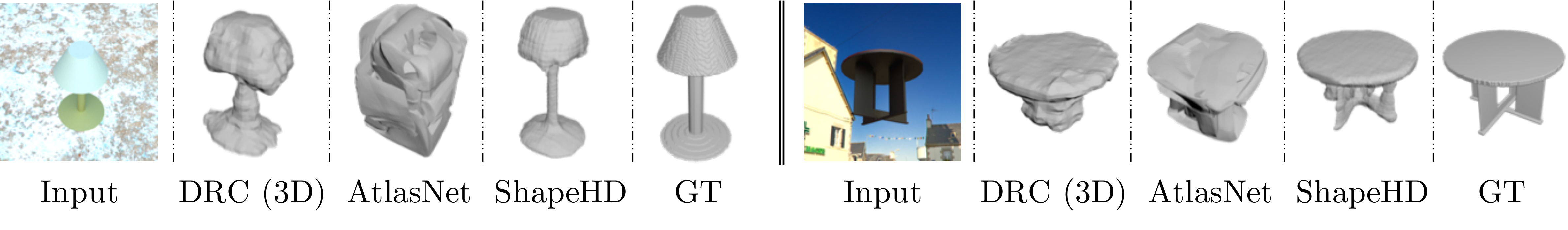}
    \scriptsize
 	\setlength{\tabcolsep}{1.4pt}
    \begin{tabular}{lcccccccccccccc}
    \toprule
    Methods & bench & boat & cabin & car & chair & disp & lamp & phone & plane & rifle & sofa & speak & table & avg \\
    \midrule
    DRC (3D)~\cite{Tulsiani2017Multi} & .122 & .131 & .127 & .077 & .128 & .128 & .168 & .102 & .166 & .107 & .106 & {\bf .138} & .138 & .126\\
    AtlasNet~\cite{groueix2017}$^*$ & .123 & .130 & .169 & .107 & .141 & .162 & .171 & .138 & .105 & .096 & .131 & .172 & .161 & .139 \\
    \model (ours) & {\bf .121} & {\bf .103} & {\bf .126} & {\bf .066} & {\bf .125} & {\bf .124} & {\bf .157} & {\bf .084} & {\bf .073} & {\bf .053} & {\bf .102} & .141 & {\bf .124} & {\bf .108}\\
    \bottomrule
    \end{tabular}
    \vspace{-5pt}
    \caption{Qualitative results and CDs for 3D shape reconstruction on ShapeNet~\cite{Chang2015Shapenet:}. Our rendering of ShapeNet is more challenging than that from Choy~\etal~\cite{Choy20163d}; as such, the numbers of the other methods may differ from those in the original paper. All methods are trained with full 3D supervision on our rendering of the largest 13 ShapeNet categories. $^*$DRC and \model take a single image as input, while AltasNet requires ground truth object silhouettes as additional input.}
    \vspace{-15pt}
    \label{fig:results_recon_shapenet}
\end{figure}

\section{3D Shape Reconstruction}
\label{sec:exp_recon}

We now evaluate \model on 3D shape reconstruction from a single color image.

\myparagraph{RGB image preparation. } 
For the task of single-image 3D reconstruction, we need to render RGB images that correspond to the depth images for training. We follow the same camera setup specified earlier. Additionally, to boost the realism of the rendered RGB images, we put three different types of backgrounds behind the object during rendering. One third of the images are rendered in a clean white background; one third are rendered in high-dynamic-range backgrounds with illumination channels that produce realistic lighting. We render the remaining one third images with backgrounds randomly sampled from the SUN database~\cite{Xiao2010Sun}. 

\myparagraph{Baselines. }
We compare our \model with the state-of-the-art in 3D shape reconstruction, including 3D-R2N2~\cite{Choy20163d}, point set generation network (PSGN)~\cite{Fan2017point}, differentiable ray consistency (DRC)~\cite{Tulsiani2017Multi}, octree generating network (OGN)~\cite{Tatarchenko2017Octree}, and AtlasNet~\cite{groueix2017}. 3D-R2N2, DRC, OGN, and our \model take a single image as input, while PSGN and AltasNet require object silhouettes as additional input. 

\begin{figure}[t!]
    \centering
    \includegraphics[width=\linewidth]{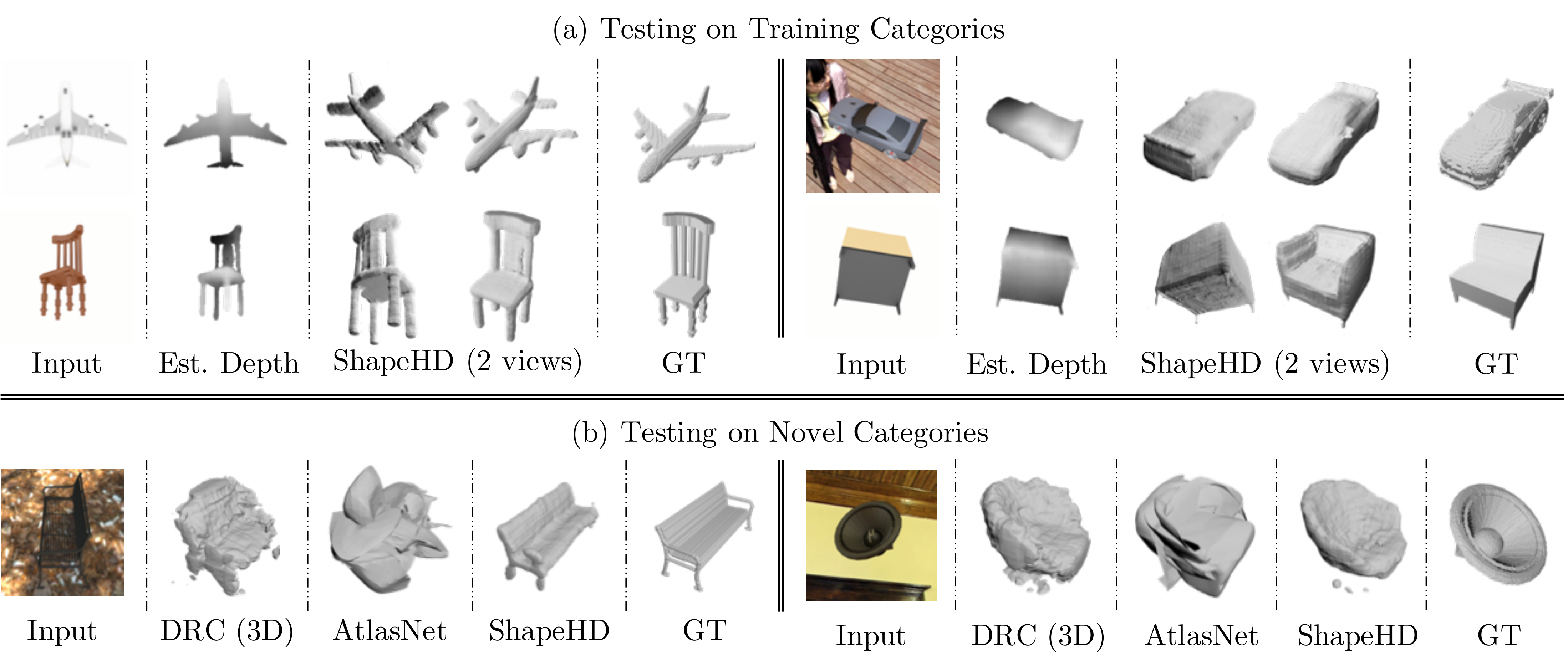}
 	\setlength{\tabcolsep}{1.3pt}
    \begin{tabular}{lccccccccccc}
    \toprule
    Methods & bench & boat & cabin & disp & lamp & phone & rifle & sofa & speak & table & avg \\
    \midrule
    DRC (3D)~\cite{Tulsiani2017Multi} & .175 & .161 & .189 & .278 & {\bf .225} & .268 & .153 & .149 & .203 & .221 & .202\\
    AtlasNet~\cite{groueix2017}$^*$ & {\bf .155} & {\bf .114} & .202 & {\bf .244} & .261 & .263 & {\bf .121} & {\bf .126} & .206 & .262 & {\bf .195} \\
    \model (ours) & .166 & .129 & {\bf .182} & .252 & .235 & {\bf .229} & .232 & .133 & {\bf .193} & {\bf .199} & {\bf .195} \\
    \bottomrule
    \end{tabular}
    \vspace{-5pt}
    \caption{Qualitative results and CDs for 3D shape reconstruction on novel categories from ShapeNet~\cite{Chang2015Shapenet:}. All methods are trained with full 3D supervision on our rendering of ShapeNet cars, chairs, and planes, and tested on the next 10 largest categories. $^*$DRC and \model take a single image as input, while AltasNet requires ground truth object silhouettes as additional input.}
    \vspace{-25pt}
    \label{fig:results_generalize_shapenet}
\end{figure}

\begin{figure}[t!]
\centering
\includegraphics[width=\linewidth]{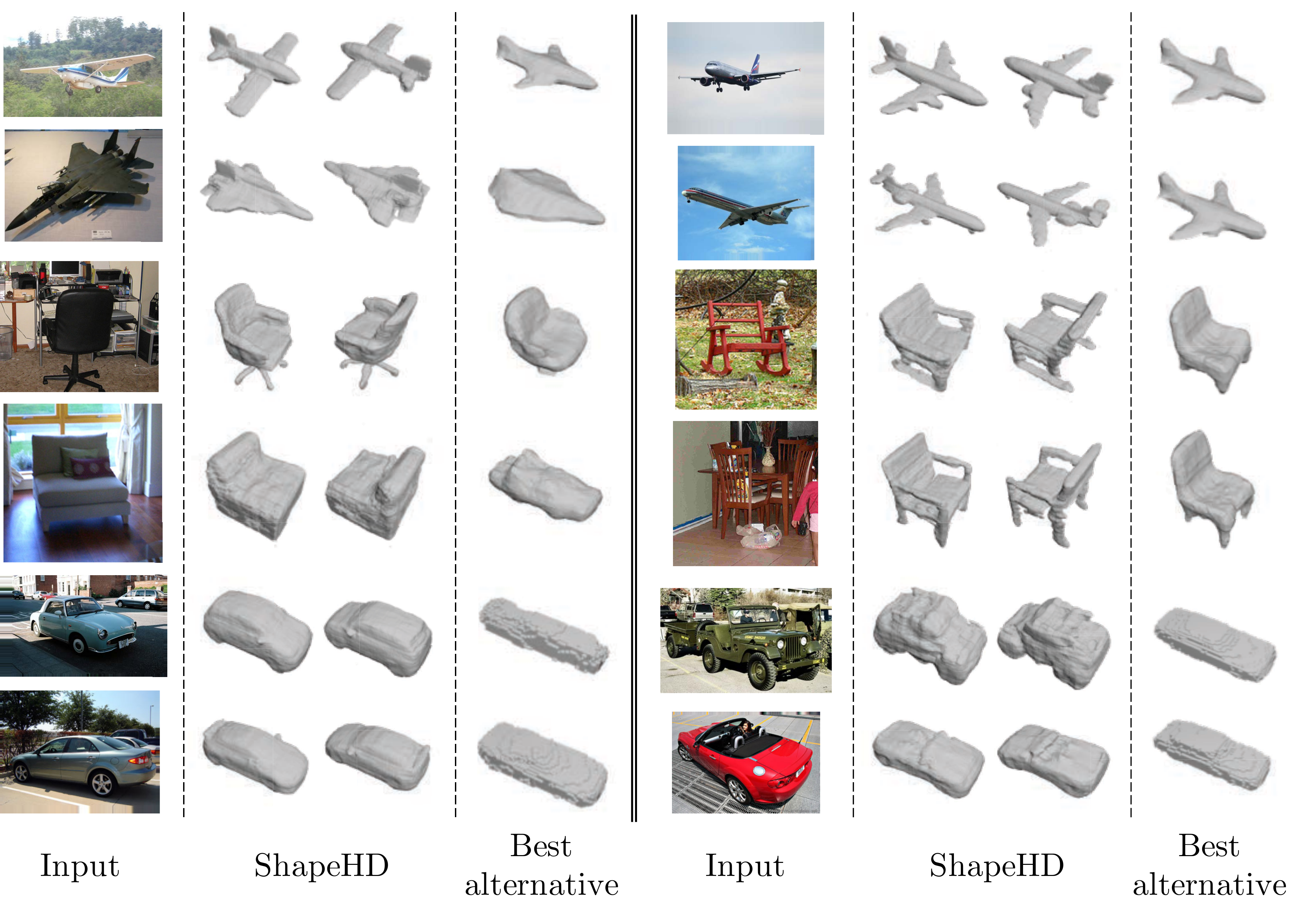}
\vspace{-25pt}
\caption{Single-view 3D shape reconstruction on PASCAL 3D+~\cite{Xiang2014PASCAL:}. From left to right: input, two views of reconstructions from \model, and reconstructions by the best alternative in~\tbl{tbl:recon_pascal}. Assisted by the learned naturalness losses, \model recovers accurate 3D shapes with fine details.}
\label{fig:results_recon_pascal}
\vspace{-8pt}
\end{figure}
\begin{figure}[t!]
\centering
\includegraphics[width=\linewidth]{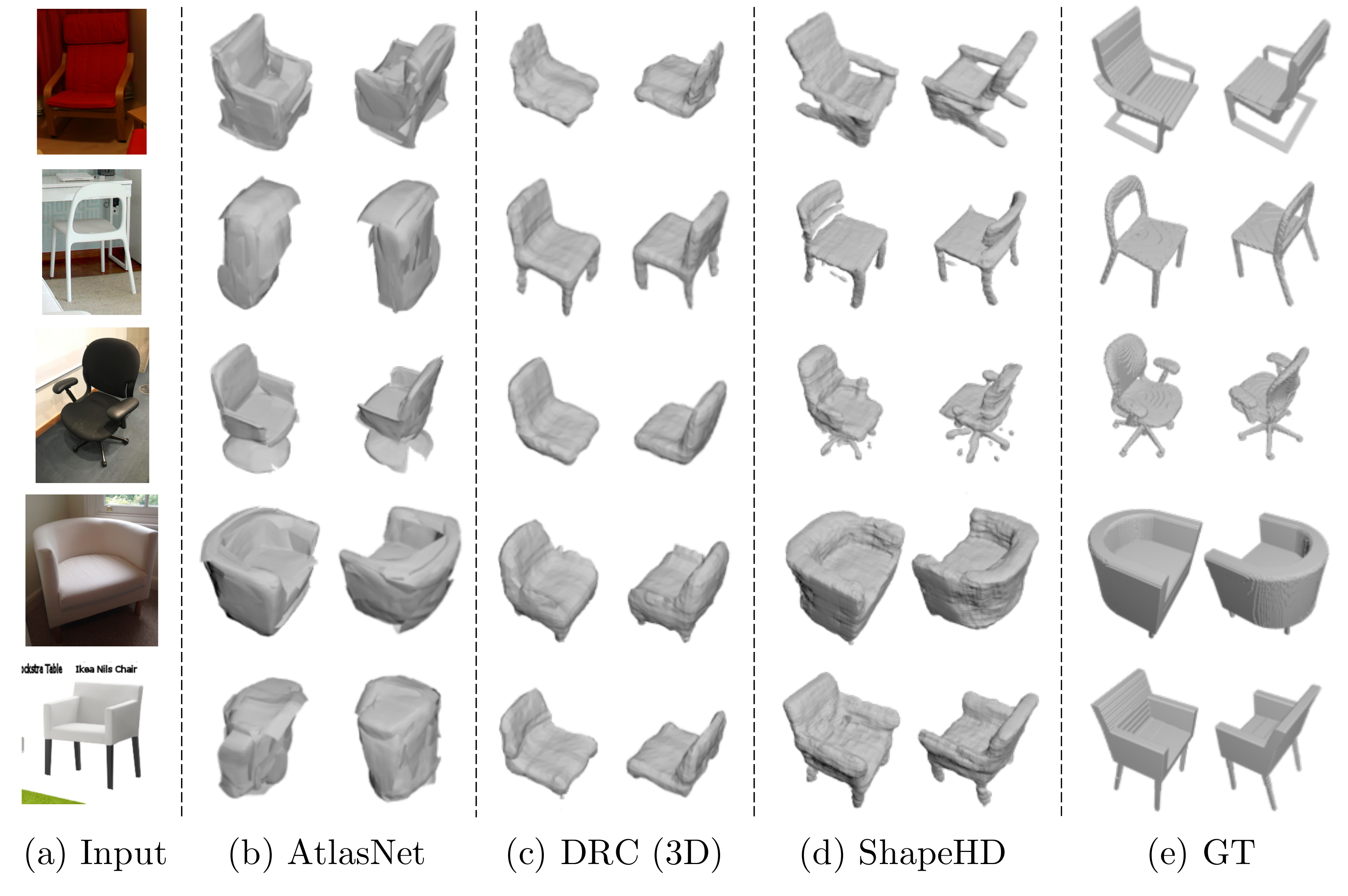}
\vspace{-25pt}
\caption{Single-view 3D reconstruction on Pix3D~\cite{pix3d}. For each input image, we show reconstructions by AtlasNet, DRC, our \model, and ground truth. Our \model reconstructs complete 3D shapes with fine details that resemble the ground truth. }
\label{fig:results_recon_pix3d}
\vspace{-25pt}
\end{figure}

\myparagraph{Results on synthetic data. }
We first evaluate on renderings of ShapeNet objects~\cite{Chang2015Shapenet:}. We present reconstructed 3D shapes and quantitative results in~\figs{fig:results_recon_shapenet}. All these models are trained on our rendering of the largest 13 ShapeNet categories (those have at least 1,000 models) with ground truth 3D shapes as supervision. In general, our \model is able to predict 3D shapes that closely resemble the ground truth shapes, giving fine details that make the reconstructed shapes more realistic. It also performs better quantitatively.

\myparagraph{Generalization on novel categories. }
An important aspect of evaluating shape reconstruction methods is on how well they generalize. Here we train our model and baselines on the largest three ShapeNet classes (cars, chairs, and planes), again with ground truth shapes as supervision, and test them on the next largest ten. \fig{fig:results_generalize_shapenet} shows our \model performs better than DRC (3D) and is comparable to AtlasNet; however, note that AtlasNet requires ground truth silhouettes as additional input, while \model works on raw images.

\begin{table}[t!]
 	\centering
    \small
    \begin{subtable}[b]{.52\linewidth}
 	\setlength{\tabcolsep}{2pt}
    \begin{tabular}{lcccc}
    \toprule
    \multirow{2}{*}{Methods} &  \multicolumn{4}{c}{CD} \\
    \cmidrule(l){2-5}
    & chair & car & plane & avg \\
    \midrule
    3D-R2N2~\cite{Choy20163d} & 0.238 & 0.305 & 0.305 & 0.284 \\
    DRC (3D)~\cite{Tulsiani2017Multi} & 0.158 & 0.099 & 0.112 & 0.122 \\
    OGN~\cite{Tatarchenko2017Octree} & - & {\bf 0.087} & - & - \\
    \model (ours) & {\bf 0.137} & 0.129 & {\bf 0.094} & {\bf 0.119} \\
    \bottomrule
    \end{tabular}
    \caption{CDs on PASCAL 3D+~\cite{Xiang2014PASCAL:}}
    \end{subtable}
    \hfill
    \begin{subfigure}[b]{.44\linewidth}
    \includegraphics[width=\linewidth]{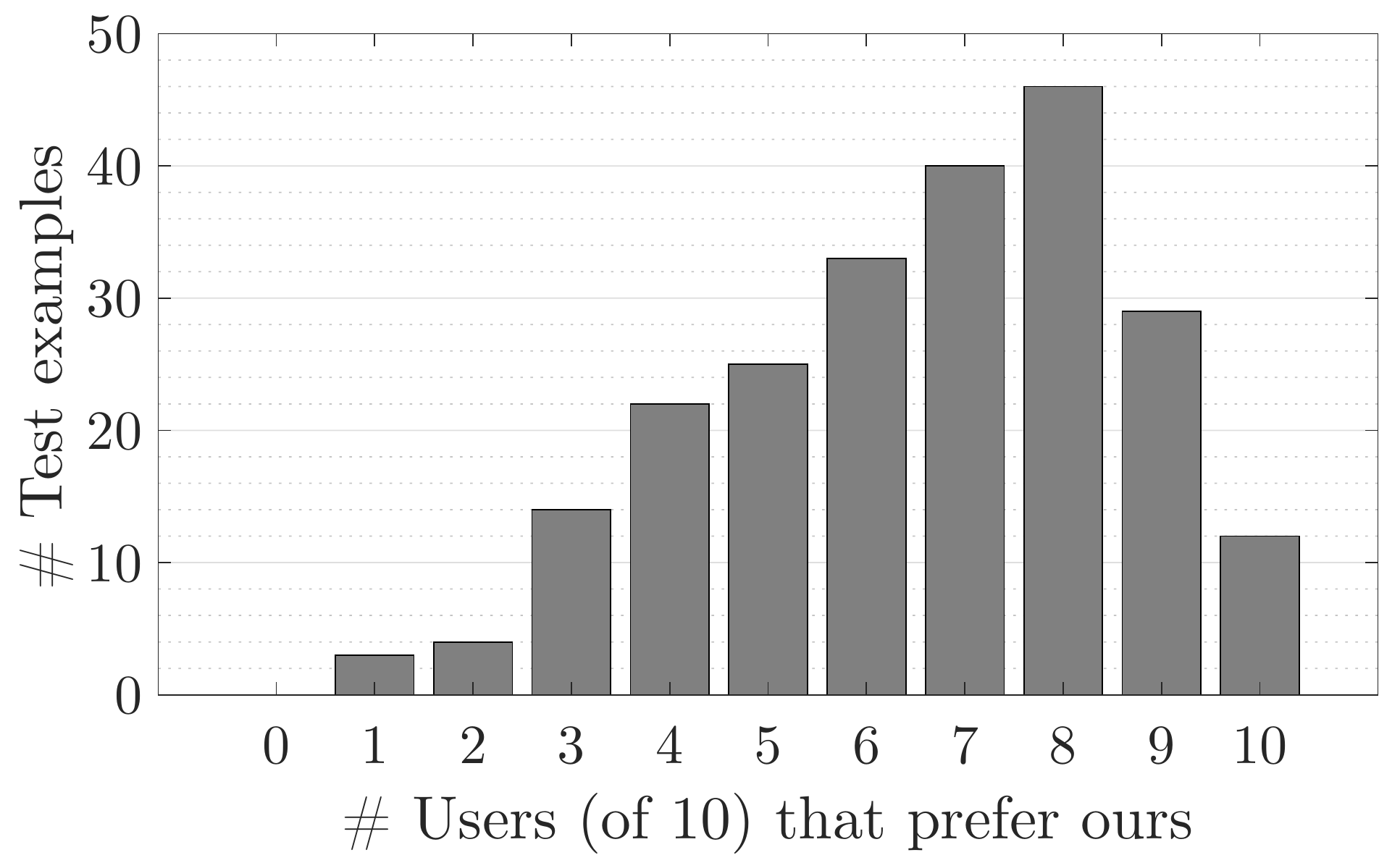}
    \vspace{-15pt}
    \caption{\label{tbl:recon_pascal_amt} Human Study results}
    \end{subfigure}
    \vspace{-7pt}
    \caption{Results for 3D shape reconstruction on PASCAL 3D+~\cite{Xiang2014PASCAL:}. (a) We compare our \model with 3D-R2N2, DRC, and OGN. PSGN and AtlasNet are not evaluated, because they require object masks as additional input, but PASCAL 3D+ has only inaccurate masks. (b) In the behavioral study, most users prefer our constructions on most images. Overall, our reconstructions are preferred 64.5\% of the time to OGN's.}
    \vspace{-30pt}
    \label{tbl:recon_pascal}
\end{table}

\myparagraph{Results on real data. }
We then evaluate on two real datasets, PASCAL 3D+~\cite{Xiang2014PASCAL:} and Pix3D~\cite{pix3d}. Here, we train our model on synthetic ShapeNet renderings and use the pre-trained models released by the authors as baselines. All methods take ground truth 3D shapes as supervision during training. As shown in \figs{fig:results_recon_pascal} and \ref{fig:results_recon_pix3d}, \model works well, inferring a reasonable shape even in the presence of strong self-occlusions. In particular, in \fig{fig:results_recon_pascal}, we compare our reconstructions with the best-performing alternatives (DRC on chairs and airplanes, and AtlasNet on cars). In addition to preserving details, our model captures the shape variations of the objects, while the competitors produce similar reconstructions across instances. 

Quantitatively, \tbls{tbl:recon_pascal} and \ref{tbl:recon_pix3d} suggest that \model performs significantly better than the other methods in almost all metrics. The only exception is the CD on PASCAL 3D+ cars, where OGN performs the best. However, as PASCAL 3D+ only has around 10 CAD models for each object category as ground truth 3D shapes, the ground truth labels and the scores can be inaccurate, failing to reflect human perception~\cite{Tulsiani2017Multi}. 

We therefore conduct an additional user study, where we show an input image and its two reconstructions (from \model and from OGN, each in two views) to users on Amazon Mechanical Turk, and ask them to choose the shape that looks closer to the object in the image. For each image, we collect 10 responses from ``Masters'' (workers who have demonstrated excellence across a wide range of HITs). \tbl{tbl:recon_pascal_amt} suggests that on most images, most users prefer our reconstruction to OGN's. In general, our reconstructions are preferred 64.5\% of the time.

\begin{table}[t!]
 	\centering
 	\small
 	\setlength{\tabcolsep}{4pt}
    \begin{tabular}{lccccc}
    \toprule
    & 3D-R2N2~\cite{Choy20163d} & DRC (3D)~\cite{Tulsiani2017Multi} & PSGN~\cite{Fan2017point}$^*$ & AtlasNet~\cite{groueix2017}$^*$ & \model \\
    \midrule
    IoU ($32^3$) & 0.136 & 0.265 & - & - & {\bf 0.284} \\
    IoU ($128^3$) & 0.089 & 0.185 & - & - & {\bf 0.205} \\
    \midrule
    CD & 0.239 & 0.160 & 0.199 & 0.126 & {\bf 0.123} \\
    \bottomrule
    \end{tabular}
    \vspace{5pt}
    \caption{\label{tbl:recon_pix3d} 3D shape reconstruction results on Pix3D~\cite{pix3d}. All methods were trained with full 3D supervision on rendered images of ShapeNet objects. $^*$3D-R2N2, DRC, and \model take a single image as input, while PSGN and AtlasNet require the ground truth mask as input. Also, PSGN and AtlasNet generate surface point clouds without guaranteeing watertight meshes and therefore cannot be evaluated in IoU.}
    \vspace{-20pt}
\end{table}

\begin{figure}[t]
\centering
\includegraphics[width=\linewidth]{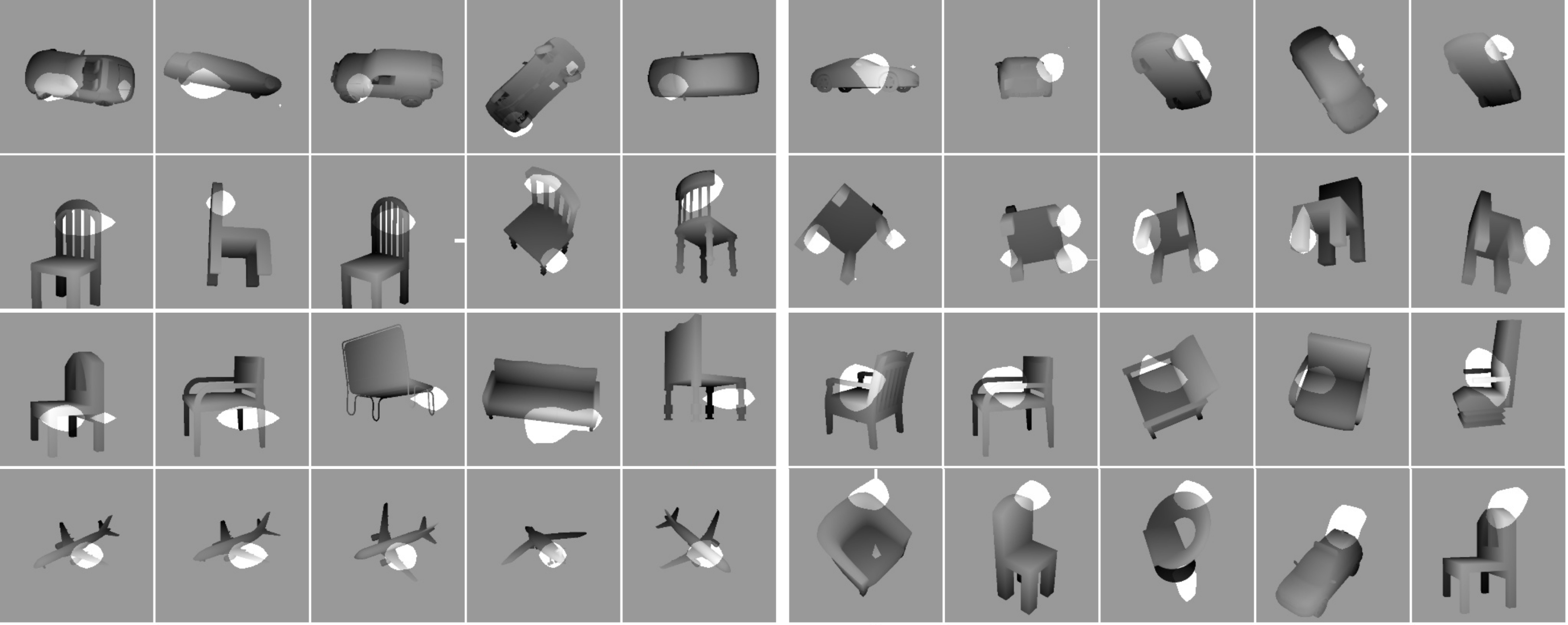}
\vspace{-20pt}
\caption{Visualizations on how \model attends to details in depth maps. Row 1: car wheel detectors. Row 2: chair back and leg detectors. The left responds to the strided pattern in particular. Row 3: chair arm and leg detectors. Row 4: airplane engine and curved surface detectors. The right responds to a specific pattern across classes.}
\label{fig:vis}
\vspace{-20pt}
\end{figure}
\section{Analyses}

We want to understand what the network has learned. In this section, we present a few analyses to visualize what the network is learning, analyze the effect of the naturalness loss function over time, and discuss common failure modes.

\myparagraph{Network visualization. }
As the network successfully reconstructs object shape and parts, it is natural to ask if it learns object or part detectors implicitly. To this end, we visualize the top activating regions across all validation images for units in the last convolutional layer of the encoder in our 3D completion network, using the method proposed by Zhou~\etal~\cite{Zhou2014Object}. As shown in \fig{fig:vis}, the network indeed learns a diverse and rich set of object and part detectors. There are detectors that attend to car wheels, chair backs, chair arms, chair legs, and airplane engines. Also note that many detectors respond to certain patterns (\eg, strided) in particular, which is probably contributing to the fine details in the reconstruction. Additionally, there are units that respond to generic shape patterns across categories, like the curve detector in the bottom right.

\begin{figure}[t!]
\centering
\includegraphics[width=\linewidth]{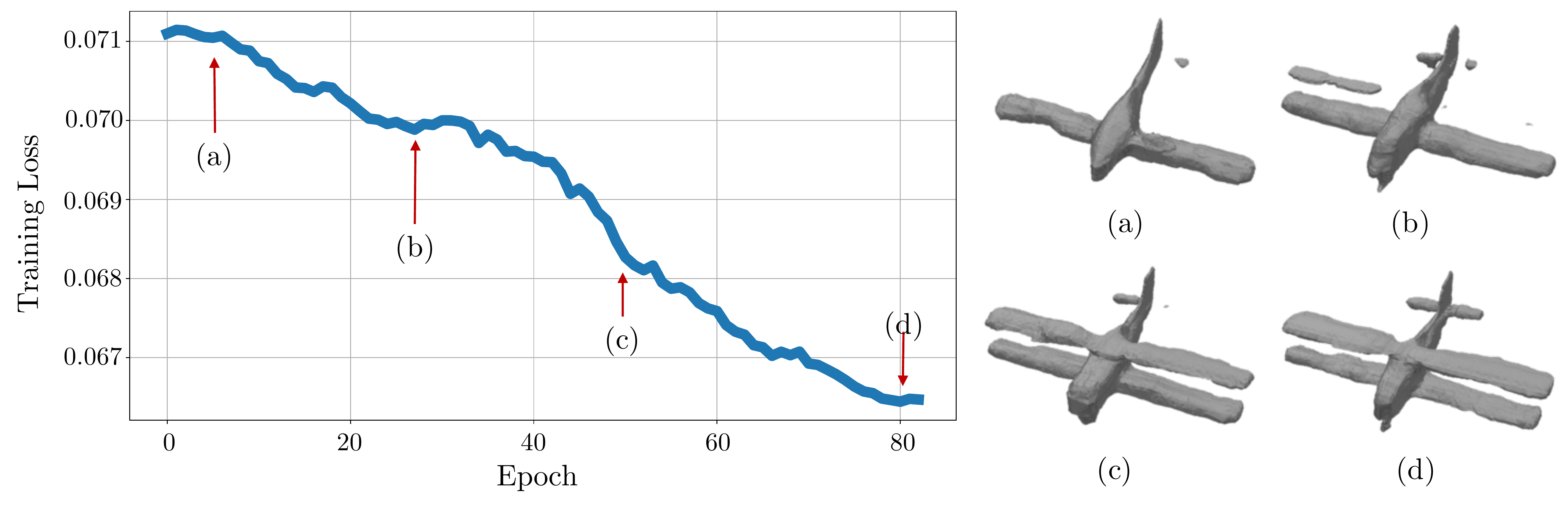}
\vspace{-25pt}
\caption{Visualizations on how \model evolves over time with naturalness losses: the predicted shape becomes increasingly realistic as details are being added.}
\label{fig:time}
\vspace{-12pt}
\end{figure}
\begin{figure}[t]
\centering
\includegraphics[width=\linewidth]{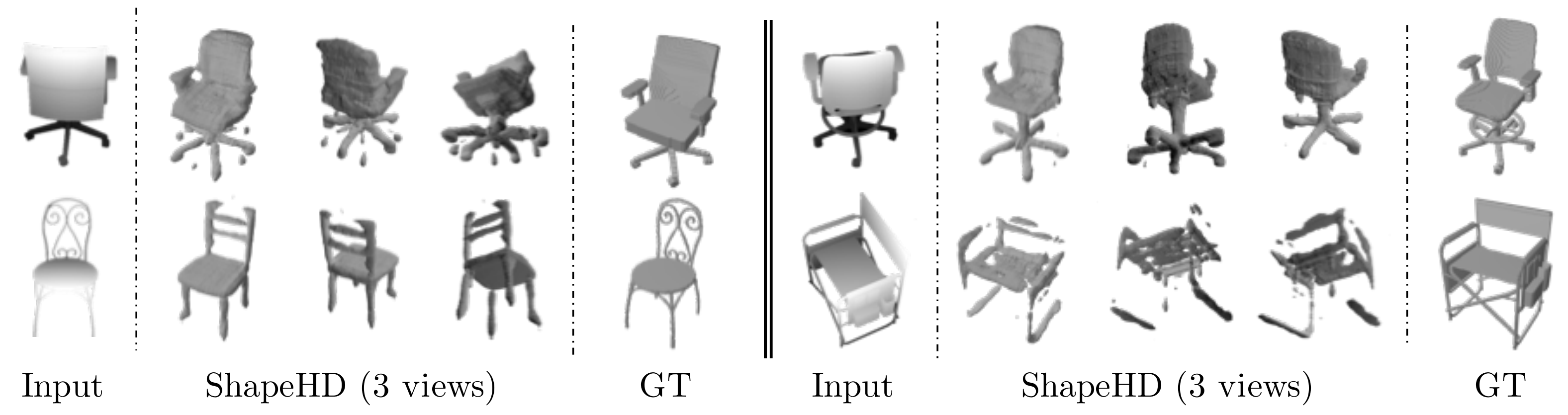}
\vspace{-20pt}
\caption{Common failure modes of our system. Top left: the model sometimes gets confused by deformable object parts (\eg, wheels). Top right: the model might miss uncommon object parts (the ring above the wheels). Bottom row: the model has difficulty in recovering very thin structure, and may generate other structure patterns instead.}
\vspace{-20pt}
\label{fig:fail}
\end{figure}

\myparagraph{Training with naturalness loss over time. }
We study the effect of the naturalness loss over time. In \fig{fig:time}, we plot the loss of the completion network with respect to fine-tuning epochs. We realize the voxel loss goes down slowly but consistently. If we visualize the reconstructed examples at different timestamps, we clearly see details are being added to the shapes. These fine details occupy a small region in the voxel grid, and thus training with supervised loss alone is unlikely to recover them. In contrast, with adversarially training perceptual losses, our model recovers details successfully. 

\myparagraph{Failure cases. }
We present failure cases in \fig{fig:fail}. We observe our model has these common failing modes: it sometimes gets confused by deformable object parts (\eg, wheels on the top left); it may miss uncommon object parts (top right, the ring above the wheels); it has difficulty in recovering very thin structure (bottom right), and may generate other patterns instead (bottom left). While the voxel representation makes it possible to incorporate the naturalness loss, intuitively, it also encourages the network to focus on thicker shape parts, as they carry more weights in the loss function. 

\section{Conclusion}

We have proposed to use learned shape priors to overcome the 2D-3D ambiguity and to learn from the multiple hypotheses that explain a single-view observation. Our \model achieves state-of-the-art results on 3D shape completion and reconstruction. We hope our results will inspire further research in 3D shape modeling, in particular on explaining the ambiguity behind partial observations. 

\vspace{2pt}
\myparagraph{Acknowledgements: } 
This work is supported by NSF \#1231216, ONR MURI N00014-16-1-2007, Toyota Research Institute, Shell Research, and Facebook.

%
%
%
\bibliographystyle{splncs04}
\bibliography{3dpercept}

\end{document}